\documentclass[letterpaper, 10 pt, conference]{ieeeconf}
\IEEEoverridecommandlockouts                              
\usepackage{amsmath}

\usepackage{amsfonts}

\usepackage{times}
\usepackage{epsfig}
\usepackage{graphicx}

\usepackage{graphicx}
\usepackage{epsfig}
\usepackage{times}
\usepackage{mathptmx}
\usepackage{cite}
\usepackage{color}

\usepackage{times}

\definecolor{red}{rgb}{1,0,0}
\definecolor{green}{rgb}{0,1,0}
\definecolor{blue}{rgb}{0,0,1}
\definecolor{violet}{rgb}{1,0,1}
\definecolor{cyan}{cmyk}{1,0,0,0}
\definecolor{magenta}{cmyk}{0,1,0,0}
\definecolor{yellow}{cmyk}{0,0,1,0}

\definecolor{white}{rgb}{1,1,1}

\newcommand{\CO}[1]{}

 \newcommand{\editage}[1]{}

\title{
\bf 
Compressive Self-localization Using Relative Attribute Embedding
}

\author{Yamamoto Ryogo ~~~~~ Tanaka Kanji
\thanks{Our work has been supported in part by 
JSPS KAKENHI 
Grant-in-Aid 
for Scientific Research (C) 17K00361 
and 20K12008.}
\thanks{The authors are with Graduate School of Engineering, University of Fukui, Japan. 
{\tt\small tnkknj@u-fukui.ac.jp}}}

\begin{document}

\newcommand{\FIG}[3]{
\begin{minipage}[b]{#1cm}
\begin{center}
\includegraphics[width=#1cm]{#2}\\
{\scriptsize #3}
\end{center}
\end{minipage}
}

\newcommand{\FIGU}[3]{
\begin{minipage}[b]{#1cm}
\begin{center}
\includegraphics[width=#1cm,angle=180]{#2}\\
{\scriptsize #3}
\end{center}
\end{minipage}
}

\newcommand{\FIGm}[3]{
\begin{minipage}[b]{#1cm}
\begin{center}
\includegraphics[width=#1cm]{#2}\\
{\scriptsize #3}
\end{center}
\end{minipage}
}

\newcommand{\FIGR}[3]{
\begin{minipage}[b]{#1cm}
\begin{center}
\includegraphics[angle=-90,width=#1cm]{#2}
\\
{\scriptsize #3}
\vspace*{1mm}
\end{center}
\end{minipage}
}

\newcommand{\FIGRpng}[5]{
\begin{minipage}[b]{#1cm}
\begin{center}
\includegraphics[bb=0 0 #4 #5, angle=-90,clip,width=#1cm]{#2}\vspace*{1mm}
\\
{\scriptsize #3}
\vspace*{1mm}
\end{center}
\end{minipage}
}

\newcommand{\FIGCpng}[5]{
\begin{minipage}[b]{#1cm}
\begin{center}
\includegraphics[bb=0 0 #4 #5, angle=90,clip,width=#1cm]{#2}\vspace*{1mm}
\\
{\scriptsize #3}
\vspace*{1mm}
\end{center}
\end{minipage}
}

\newcommand{\FIGpng}[5]{
\begin{minipage}[b]{#1cm}
\begin{center}
\includegraphics[bb=0 0 #4 #5, clip, width=#1cm]{#2}\vspace*{-1mm}\\
{\scriptsize #3}
\vspace*{1mm}
\end{center}
\end{minipage}
}

\newcommand{\FIGtpng}[5]{
\begin{minipage}[t]{#1cm}
\begin{center}
\includegraphics[bb=0 0 #4 #5, clip,width=#1cm]{#2}\vspace*{1mm}
\\
{\scriptsize #3}
\vspace*{1mm}
\end{center}
\end{minipage}
}

\newcommand{\FIGRt}[3]{
\begin{minipage}[t]{#1cm}
\begin{center}
\includegraphics[angle=-90,clip,width=#1cm]{#2}\vspace*{1mm}
\\
{\scriptsize #3}
\vspace*{1mm}
\end{center}
\end{minipage}
}

\newcommand{\FIGRm}[3]{
\begin{minipage}[b]{#1cm}
\begin{center}
\includegraphics[angle=-90,clip,width=#1cm]{#2}\vspace*{0mm}
\\
{\scriptsize #3}
\vspace*{1mm}
\end{center}
\end{minipage}
}

\newcommand{\FIGC}[5]{
\begin{minipage}[b]{#1cm}
\begin{center}
\includegraphics[width=#2cm,height=#3cm]{#4}~$\Longrightarrow$\vspace*{0mm}
\\
{\scriptsize #5}
\vspace*{8mm}
\end{center}
\end{minipage}
}

\newcommand{\FIGf}[3]{
\begin{minipage}[b]{#1cm}
\begin{center}
\fbox{\includegraphics[width=#1cm]{#2}}\vspace*{0.5mm}\\
{\scriptsize #3}
\end{center}
\end{minipage}
}

\newcommand{\figplan}[1]{~}

\newcommand{\SW}[2]{#1}

\newcommand{\figA}{
\begin{figure}[t]
% \FIGpng{8}{figA.png}{}{300}{100}
\FIG{8}{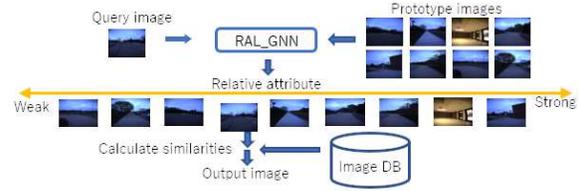}{}
\caption{Relative attribute-based embedding and its application to VPR.}\label{fig:A}
\end{figure}
}

\newcommand{\figB}{
\begin{figure}[t]
% \FIGpng{8}{NCLTfig.png}{}{370}{200}
\FIG{8}{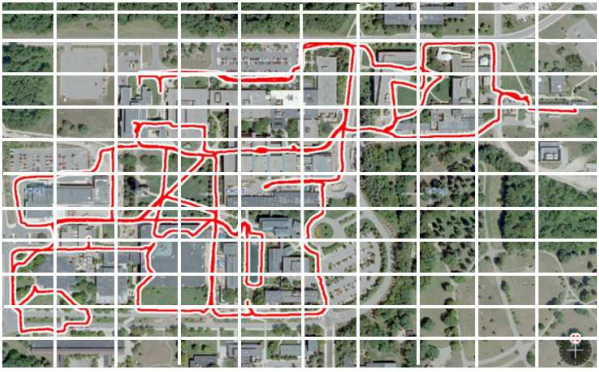}{}
\caption{The robot workspace and place classes.}\label{fig:B}
\end{figure}
}

\newcommand{\figC}{
\begin{figure}[t]
% \FIGpng{8}{nclt_prototype.png}{}{420}{250}
\FIG{8}{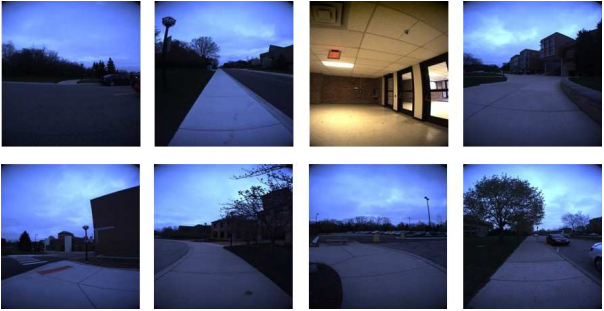}{}
\caption{Prototype images.}\label{fig:C}
\end{figure}
}

\newcommand{\figD}{
\begin{figure}[t]
% \FIGpng{8}{osr.png}{}{420}{250}
\FIG{8}{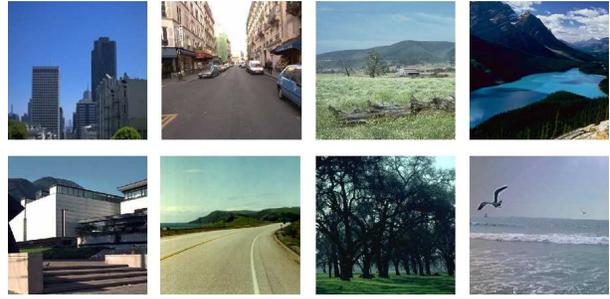}{}
\caption{Example images of 8 different categories in the OSR dataset.}\label{fig:D}
\end{figure}
}

\newcommand{\tabA}{
\begin{table}[t]
\caption{Performance results.}\label{tab:A}
\begin{tabular}{ll}
BRS & 0.301 \\
RRS & 0.426
\end{tabular}
\end{table}
}

\maketitle

\author{}

\begin{abstract}
The use of relative attribute (e.g., beautiful, safe, convenient) -based image embeddings in visual place recognition, as a domain-adaptive compact image descriptor that is orthogonal to the typical approach of absolute attribute (e.g., color, shape, texture) -based image embeddings, is explored in this paper.
\end{abstract}

\section{Introduction}

Most current state-of-the-art visual place recognition (VPR) algorithms employ absolute attribute (e.g., color, shape, texture) -based image embedding for image feature description \cite{SIFT, ref26, GIST} and image similarity search \cite{topo1}. In this study, we are interested in relative attributes (e.g., beautiful, safe, convenient) \cite{souri2016deep, sandeep2014relative, parikh2011relative}-based image embedding, as it provides a domain-adaptive ranking-based image description \cite{kanji2021deep} and it is orthogonal to typical approach of absolute attributes-based embeddings. Specifically, we present two different solutions based on binary and real-valued relative attribute strength and experimentally evaluate them via cross-season VPR experiments \cite{NCLT}.

\section{Approach}

VPR is formulated as a problem of similar image retrieval \cite{survey16vpr, zhang2021visual, garg2021your}. The objective is to search for the image most relevant to a given query image over an image database. The database is constructed as a collection of viewpoint-annotated view images from visual experiences in the training domain via structure-from-motion \cite{sfm} or SLAM \cite{slam}. Specifically, the procedure for construction consists of two steps (Fig. \ref{fig:A}): (1) extracting a feature descriptor from the image, and (2) evaluating the descriptor similarity between the query and each database images. Either step is detailed in the following.

\subsection{Feature Descriptor}

An input query/database image is described by measuring the relative attribute strength in $\mathbb{R}$, 
with respect to a predefined prototype image \cite{kanji2021deep}. 
If the strength value is negative, it means that the input image has stronger relative attribute than the prototype image, otherwise it means that it is weaker. 
(1) Specifically, the processing begins by evaluating the relative attribute strength $p_{i}^{\bf attr}$ of each $i$-th prototype image 
$p_{i}$ with respect to the input image $q$ for each $j$-th relative attribute model 
$A_{j}$, which yields a length $(N+1)$ list of relative attribute strength 
\begin{equation}
L=(q_{j}^{\bf attr}, p_{j1}^{\bf attr}, \cdots, p_{jN}^{\bf attr})
\end{equation}
with a boundary condition 
\begin{equation}
q_{j}^{\bf attr}=0. 
\end{equation}
(2) Then, the list is sorted in the descending order of relative attribute strength, which yields an ordered list $L'$.
(3) Then, each image is ranked based on the ranking $L'$, which yields a 1$\times$$(N+1)$ matrix $R_{j}$ of rank:
\begin{equation}
(q_{j}^{\bf rank}, p_{j1}^{\bf rank}, \cdots, p_{jN}^{\bf rank}). 
\end{equation}
(4) By performing the above processes for a predefined set of $M$ relative attributes, we obtain an $M{\times}(N+1)$ matrix $R$ as the final output of the feature descriptor step.

\subsection{Descriptor Similarity}

Next, descriptor similarity is evaluated between the input descriptor 
$R$ 
and each database descriptor 
$R'$, 
for which we have developed two different kinds of evaluation methods. 
The first method, called binary relative strength (BRS), treats the descriptor as a binary relative attribute (stronger or weaker), and evaluates the similarity by 
\begin{equation}
BRS= 
\sum_j | R_{j1}-R'_{j1} |. 
\end{equation}
The second method,  called ranked relative strength (RRS), utilizes the real-valued strength from the descriptor, and evaluates the similarity by 
\begin{equation}
RRS=
\sum_i \sum_j | R_{ji}-R'_{ji} |.
\end{equation}

\figA

\section{EXPERIMENTS}

\figB

% \tabA

\figC

\figD

The experimental settings follow the procedure in \cite{DBLP:journals/corr/abs-2109-04569}.
The NCLT dataset used is a large scale, long-term autonomy dataset collected by a Segway robot in a university campus.
Specifically,
view images from the on-board front-facing camera (Ladybug camera) in the sessions 2012/03/31 and 2012/08/04 were used as training and test image sets, respectively. 
We considered a place classification task with a set of 8 place classes.
Specifically, the entire workspace of the range 
$[-740, 130]{\times}[-330, 120]$ 
was divided 
via grid-based place partitioning
into a 
10$\times$10 grid 
of
100 place classes (Fig. \ref{fig:B}),
from which the 8 classes are randomly sampled.
We simply sample
$N=8$ prototype images 
from each of the 8 place classes (Fig. \ref{fig:C}). 
The relative attribute models 
were trained using the OSR dataset 
as in 
\cite{meng2018efficient} (Fig. \ref{fig:D})
on the $M=6$ different attribute classes: natural, open, perspective, large-objects, diagonal-plane and close-depth.
% Figure ? shows the performance results for the two types of similar image retrieval methods described in Section ?. 
The training, testing and performance evaluation 
were iterated for 100 sets of randomly sampled 8 place classes.
The mAP performance was 
0.341$\pm$0.071
and
0.364$\pm$0.079
for the BRS and RRS methods, respectively.
For comparison, 
absolute attribute counterparts of 
BRS and RRS were also developed 
using the 1-hot semantic histogram as the absolute attribute feature as in \cite{DBLP:journals/corr/abs-2109-04569},
and tested with results of 
0.147$\pm$0.068 and 
0.310$\pm$0.090, 
respectively.
One can see that the RRS method outperforms the BRS method in the current experiments. Specifically, the real-valued relative attribute strength provided rich information, and the information loss was significant when it was binarized (i.e., the BRS method).

\section{Conclusions and Future Workds}

The use of relative attribute (e.g., beautiful, safe, convenient) -based image embeddings in visual place recognition, as a domain-adaptive compact image descriptor that is orthogonal to the typical approach of absolute attribute (e.g., color, shape, texture) -based image embeddings, is explored in this paper.
In the future, we plan to integrate the proposed highly-efficient VPR method to train a visual navigation system as in \cite{DBLP:journals/corr/abs-2109-04569}. During the training phase, the VPR module must be repeated a very large number of times. That is, the robot needs to experience a large number of (e.g., tens of thousands of) training episodes, and each training episode involves performing VPR at many (e.g., 10) viewpoints. 
Towards that goal, further acceleration of the VPR module while retraining the discriminative power is desired.

\bibliographystyle{IEEEtran}
\bibliography{ref}

\begin{thebibliography}{10}
\providecommand{\url}[1]{#1}
\csname url@rmstyle\endcsname
\providecommand{\newblock}{\relax}
\providecommand{\bibinfo}[2]{#2}
\providecommand\BIBentrySTDinterwordspacing{\spaceskip=0pt\relax}
\providecommand\BIBentryALTinterwordstretchfactor{4}
\providecommand\BIBentryALTinterwordspacing{\spaceskip=\fontdimen2\font plus
\BIBentryALTinterwordstretchfactor\fontdimen3\font minus
  \fontdimen4\font\relax}
\providecommand\BIBforeignlanguage[2]{{%
\expandafter\ifx\csname l@#1\endcsname\relax
\typeout{** WARNING: IEEEtran.bst: No hyphenation pattern has been}%
\typeout{** loaded for the language `#1'. Using the pattern for}%
\typeout{** the default language instead.}%
\else
\language=\csname l@#1\endcsname
\fi
#2}}

\bibitem{SIFT}
D.~G. Lowe, ``Distinctive image features from scale-invariant keypoints,''
  \emph{International journal of computer vision}, vol.~60, no.~2, pp. 91--110,
  2004.

\bibitem{ref26}
N.~S{\"{u}}nderhauf, S.~Shirazi, F.~Dayoub, B.~Upcroft, and M.~Milford, ``On
  the performance of convnet features for place recognition,'' in \emph{2015
  {IEEE/RSJ} International Conference on Intelligent Robots and Systems, {IROS}
  2015, Hamburg, Germany, September 28 - October 2, 2015}, 2015, pp.
  4297--4304.

\bibitem{GIST}
A.~Oliva and A.~Torralba, ``Modeling the shape of the scene: A holistic
  representation of the spatial envelope,'' \emph{International journal of
  computer vision}, vol.~42, no.~3, pp. 145--175, 2001.

\bibitem{topo1}
M.~Cummins and P.~M. Newman, ``Appearance-only {SLAM} at large scale with
  {FAB-MAP} 2.0,'' \emph{I. J. Robotics Res.}, vol.~30, no.~9, pp. 1100--1123,
  2011.

\bibitem{souri2016deep}
Y.~Souri, E.~Noury, and E.~Adeli, ``Deep relative attributes,'' in \emph{Asian
  conference on computer vision}.\hskip 1em plus 0.5em minus 0.4em\relax
  Springer, 2016, pp. 118--133.

\bibitem{sandeep2014relative}
R.~N. Sandeep, Y.~Verma, and C.~Jawahar, ``Relative parts: Distinctive parts
  for learning relative attributes,'' in \emph{Proceedings of the IEEE
  Conference on Computer Vision and Pattern Recognition}, 2014, pp. 3614--3621.

\bibitem{parikh2011relative}
D.~Parikh and K.~Grauman, ``Relative attributes,'' in \emph{2011 International
  Conference on Computer Vision}.\hskip 1em plus 0.5em minus 0.4em\relax IEEE,
  2011, pp. 503--510.

\bibitem{kanji2021deep}
K.~Tanaka, ``Deep simbad: Active landmark-based self-localization using
  ranking-based scene descriptor,'' \emph{arXiv preprint arXiv:2109.02786},
  2021.

\bibitem{NCLT}
N.~Carlevaris-Bianco, A.~K. Ushani, and R.~M. Eustice, ``University of michigan
  north campus long-term vision and lidar dataset,'' \emph{The International
  Journal of Robotics Research}, vol.~35, no.~9, pp. 1023--1035, 2016.

\bibitem{survey16vpr}
S.~M. Lowry, N.~S{\"{u}}nderhauf, P.~Newman, J.~J. Leonard, D.~D. Cox, P.~I.
  Corke, and M.~J. Milford, ``Visual place recognition: {A} survey,''
  \emph{{IEEE} Trans. Robotics}, vol.~32, no.~1, pp. 1--19, 2016.

\bibitem{zhang2021visual}
X.~Zhang, L.~Wang, and Y.~Su, ``Visual place recognition: A survey from deep
  learning perspective,'' \emph{Pattern Recognition}, vol. 113, p. 107760,
  2021.

\bibitem{garg2021your}
S.~Garg, T.~Fischer, and M.~Milford, ``Where is your place, visual place
  recognition?'' \emph{arXiv preprint arXiv:2103.06443}, 2021.

\bibitem{sfm}
J.~L. Schonberger and J.-M. Frahm, ``Structure-from-motion revisited,'' in
  \emph{Proceedings of the IEEE conference on computer vision and pattern
  recognition}, 2016, pp. 4104--4113.

\bibitem{slam}
J.~Engel, T.~Sch{\"o}ps, and D.~Cremers, ``Lsd-slam: Large-scale direct
  monocular slam,'' in \emph{European conference on computer vision}.\hskip 1em
  plus 0.5em minus 0.4em\relax Springer, 2014, pp. 834--849.

\bibitem{DBLP:journals/corr/abs-2109-04569}
M.~Yoshida, R.~Yamamoto, and K.~Tanaka, ``{S3G-ARM:} highly compressive visual
  self-localization from sequential semantic scene graph using absolute and
  relative measurements,'' \emph{CoRR}, 2021.

\bibitem{meng2018efficient}
Z.~Meng, N.~Adluru, H.~J. Kim, G.~Fung, and V.~Singh, ``Efficient relative
  attribute learning using graph neural networks,'' in \emph{ECCV}, 2018, pp.
  552--567.

\end{thebibliography}

\end{document}